\documentclass[10pt,twocolumn,letterpaper]{article}

\usepackage{iccv}
\usepackage{times}
\usepackage{epsfig}
\usepackage{graphicx}
\usepackage{amsmath}
\usepackage{amssymb}
\usepackage{booktabs}
\usepackage{multirow}
\usepackage{pifont}
\usepackage{subcaption}
\usepackage[pagebackref=true,breaklinks=true,colorlinks,bookmarks=false]{hyperref}

\usepackage[capitalize]{cleveref}
\crefname{section}{Sec.}{Secs.}
\Crefname{section}{Section}{Sections}
\Crefname{table}{Table}{Tables}
\crefname{table}{Tab.}{Tabs.}

\newcommand{\X}{\mathbf{X}}
\newcommand{\Y}{\mathbf{Y}}

\newcommand{\Q}{\mathbf{Q}}
\newcommand{\K}{\mathbf{K}}
\newcommand{\V}{\mathbf{V}}

\renewcommand{\Re}{\mathbb{R}}

\newcommand{\MHA}{\operatorname{MHA}}
\newcommand{\Att}{\operatorname{Att}}
\newcommand{\Transformer}{\operatorname{Transformer}}

\newcommand{\FFB}{\operatorname{FFB}}
\newcommand{\Softmax}{\operatorname{Softmax}}

\newcommand{\npar}[1]{\noindent\textbf{#1}}
\newcommand{\spar}[1]{\smallskip\noindent\textbf{#1}}

\newcommand{\cmark}{\ding{51}}%

\iccvfinalcopy 

\def\iccvPaperID{5644} 

\ificcvfinal\fi

\begin{document}

\title{Exploring the Role of Audio in Video Captioning}

\author{Yuhan Shen$^{\dagger}$ \thanks{Work done during YS's internship at ByteDance.}, Linjie Yang$^{\ddagger}$, Longyin Wen$^{\ddagger}$, Haichao Yu$^{\ddagger}$, Ehsan Elhamifar$^{\dagger}$, Heng Wang$^{\ddagger}$ \\
$^{\dagger}$ Northeastern University \quad \quad \quad $^{\ddagger}$ ByteDance \\
{\tt\small  \{shen.yuh,e.elhamifar\}@northeastern.edu} \\
{\tt\small  \{linjie.yang,longyin.wen,haichaoyu,heng.wang\}@bytedance.com}}

\maketitle

\begin{abstract}
   Recent focus in video captioning has been on designing architectures that can consume both video and text modalities, and using large-scale video datasets with text transcripts for pre-training, such as HowTo100M. Though these approaches have achieved significant improvement, the audio modality is often ignored in video captioning. In this work, we present an audio-visual framework, which aims to fully exploit the potential of the audio modality for captioning. Instead of relying on text transcripts extracted via automatic speech recognition (ASR), we argue that learning with raw audio signals can be more beneficial, as audio has additional information including acoustic events, speaker identity, \etc. Our contributions are twofold. First, we observed that the model {overspecializes} to the audio modality when pre-training with both video and audio modality,
   since the ground truth (\ie, text transcripts) can be solely predicted using audio. We proposed a Modality Balanced Pre-training (MBP) loss to mitigate this issue and significantly improve the performance on downstream tasks. Second, we slice and dice different design choices of the cross-modal module, which may become an information bottleneck and generate inferior results. We proposed new local-global fusion mechanisms to improve information exchange across audio and video. We demonstrate significant improvements by leveraging the audio modality on four datasets, and even outperform the state of the art on some metrics without relying on the text modality as the input.
\end{abstract}

\section{Introduction}
\label{sec:intro}

Large-scale pre-training~\cite{akbari2021vatt,kenton2019bert,miech2020end,radford2021learning,sun2019videobert,wang2022git,zellers2022merlot}
plays a key role in boosting the performance of modern deep learning models. It is even more so for vision and language tasks, such as video captioning\cite{aafaq2019spatio,gao2017video,pan2020spatio,pei2019memory,wang2018reconstruction,yan2019stat},
where leveraging large video datasets with text supervision for pre-training is essential to achieve competitive results. However, manually annotating captions for video datasets is costly and not scalable. Thus existing video captioning datasets~\cite{shi2019dense,wang2019vatex,xu2016msr,zhou2018towards}
are often limited in size. 

\begin{figure*}[!t]
    \centering
    \includegraphics[width=0.95\linewidth]{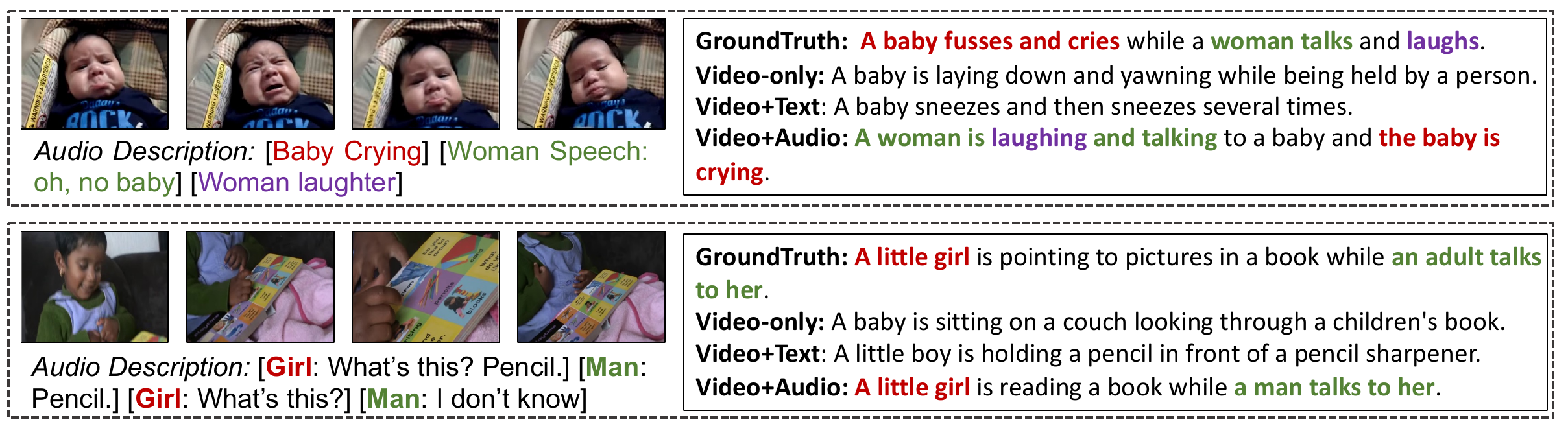}
    \caption[]{Audio provides critical complementary
    information in multi-modal video captioning. We show two examples of generated captions when we input 1) only video, 2) video and ASR text, and 3) video and audio. Audio can provide additional information that cannot be obtained from visual modality or ASR text, \eg, sound of crying, laughter, and speaker gender.\footnotemark} 
    \label{fig:pipeline-new}
\end{figure*}

To address this challenge, recent work collected datasets from instructional videos, where ASR generated transcripts can be used as text supervision, \eg, How2 \cite{sanabria2018how2}, CrossTask \cite{zhukov2019cross}, HowTo100M \cite{miech2019howto100m}, HD-VILA-100M \cite{xue2022advancing}, \etc. 
This has established a new trend of pre-training on large-scale video datasets with text transcripts for video captioning~\cite{huang2020multimodal,Luo2020UniVL,seo2022end}. We argue that text transcripts from ASR only includes partial information from audio, and hypothesize end-to-end learning using the audio modality can potentially lead to better performance, since audio can provide additional information (shown in Fig.~\ref{fig:pipeline-new}) including acoustic events, speaker identity, \etc.

\begin{figure*}[!t]
    \centering
    \includegraphics[width=0.99\linewidth]{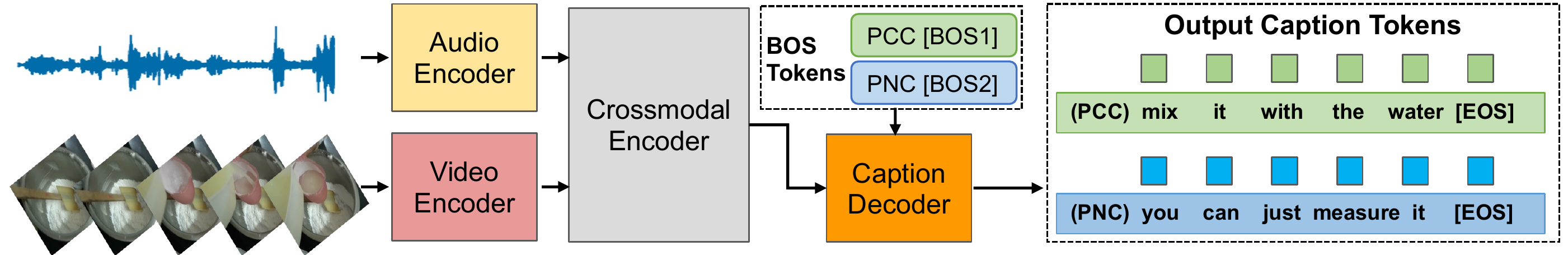}
    \caption {Overview of {our audio-visual video captioning framework}. We design two tasks for caption generation during pre-training:  Predict Current Caption (PCC) and Predict Next Caption (PNC). For downstream fine-tuning, we only adopt PCC because the goal is to predict current caption given the input frames and audio. 
    }
    \label{fig:framework}
\end{figure*}

More specifically, our paper seeks to better understand the following questions:
\vspace{-2mm}
\begin{itemize}
\item\emph{To what extent, can the audio modality improve video captioning?}
\vspace{-2mm}
\item \emph{How can the potential of the audio modality be fully realized in an audio-visual framework for captioning?}
\end{itemize}
\vspace{-2mm}
To this end, we start with a simple multi-modal pre-training framework for video captioning with ASR transcripts as supervision (shown in Fig.~\ref{fig:framework}), and look into different components that may hinder the performance of the pre-trained audio-visual model on the downstream datasets.

First, we observed that simply jointly training of the audio and video modalities may result in {degenerated models that overspecialize} to the audio modality and underfit on the video modality. As text transcripts are used as video captions during pre-training, the model essentially learns to cheat and solve the ASR problem instead of extracting information from both visual and audio signals. 
To mitigate this issue, we proposed the Modality Balanced Pre-training (MBP)
loss, which takes into account both the unimodal losses and cross-modal loss. A weighting mechanism is introduced to automatically balance different modalities during training. Fig.~\ref{fig:training-loss} shows that our MBP loss enforces the model to pay more attention to the underfitted video modality and drives the final loss much smaller. 

Second, we did a thorough study on the design of the cross-modal fusion module, which is responsible for the information exchange between the audio and video modality. 
An improperly designed cross-modal fusion module may become an information bottleneck and result in inferior performance for video captioning. 
We proposed new local-global fusion mechanisms to encourage the information flow cross different modalities. We analyzed the relevance of the annotated captions to the audio modality on downstream datasets, and observed that the local fusion mechanisms are more beneficial
to the flow of fine-grained information like single words in speech, while the global fusion mechanisms are more effective on holistic information like acoustic events or scenes. The local-global design enables the fusion module to capture information at different granularities, and mingle audio and video information at different levels. Compared with existing designs, our local-global fusion has shown empirically better results. 

By combining the two contributions, we demonstrate that audio is crucial to video captioning and provides both speech and non-speech information. Fig.~\ref{fig:pipeline-new} shows a few examples on how our model effectively integrates the information from both the audio and video modality, and generates better captions than video-only and video-text variants. 

We summarize our contributions as follows:
\vspace{-2mm}
\begin{itemize}
\setlength\itemsep{0em}
    \item Proposed to pre-train video captioning models based on video and audio modalities, explored the role of audio in video captioning, and demonstrated the benefits of audio on four benchmarks.
    \item Proposed the MBP loss to balance different modalities automatically during training, and ease the issue of overspecialization to the audio modality.
    \item Did an extensive evaluation on the effects of different cross-modal fusion modules on audio-visual video captioning, and proposed a novel local-global fusion module to effectively integrate audio and video information for video captioning.
\end{itemize}

\footnotetext{ YouTube ID of the two videos: \href{https://www.youtube.com/watch?v=a50yimv7Lqg}{a50yimv7Lqg} and \href{https://www.youtube.com/watch?v=BFYBAg9fSL4}{BFYBAg9fSL4}.}

\vspace{-2mm}
\section{Related Work}
\label{sec:related-work}
\vspace{-2mm}
\npar{Video Captioning.} Most works in video captioning \cite{aafaq2019spatio,gao2017video,pan2020spatio,pei2019memory,wang2018reconstruction,yan2019stat} focus on designing a better model to generate text descriptions given precomputed video features via an encoder-decoder framework. SwinBert \cite{lin2022swinbert} attempted to train the encoder-decoder framework directly from raw video pixels instead of refining precomputed features. In addition to visual modality, some works studied video captioning from visual data and ASR texts \cite{hessel2019case,Luo2020UniVL,seo2022end,seo2021look,shi2019dense,tang2021decembert}. A few prior works also studied audio-visual video captioning \cite{chen2019motion,BMT_Iashin_2020,rahman2019watch,tian2019audio}, but they are often limited to small-scale video captioning datasets and  precomputed input features. To the best of our knowledge, we propose the first end-to-end audio-visual video captioning framework. 

\npar{Multi-Modal Pre-training.} A growing number of works are investigating multi-modal pre-training in videos, \eg, video-text pre-training \cite{lei2021less,Luo2020UniVL,miech2020end,miech2019howto100m,patrick2020support,sun2019videobert,zellers2021merlot} and video-text-audio pre-training \cite{akbari2021vatt,alayrac2020self,zellers2022merlot}, which mostly adopt contrastive learning and/or masked language modeling to learn better representations for downstream tasks.
As only encoders are trained for multiple modalities, a separate decoder needs to be trained on top of the encoders for generative tasks such as video captioning. A recent work MV-GPT \cite{seo2022end} shows the benefits of pre-training an end-to-end encoder-decoder framework to video captioning. Unlike MV-GPT that relies on ASR text as input, our framework directly uses video and audio. A Textless Vision-Language Transformer (TVLT) \cite{tang2022tvlt} was recently proposed to take visual and audio inputs for multi-modal representation learning without ASR inputs. However, the pre-trained TVLT is a discriminative model that cannot be directly applied to generative tasks. 
While a multi-modal network receives more information and is expected to boost performance, recent works \cite{jaegle2021perceiver,panda2021adamml,peng2022balanced,wang2020makes} have identified a key challenge in training a multi-modal network that one modality may converge faster than other modalities and undermine the representation learning  of other modalities. We propose a Modality Balanced Pre-training objective to mitigate this issue and facilitate a powerful audio-visual video captioning model.

\npar{Cross-Modal Fusion.}  Given the representations of multiple modalities, a cross-modal fusion module \cite{jia2021scaling,miech2020end,miech2019howto100m,radford2021learning} will fuse these
representations into a shared embedding space to generate cross-modal representations. 
In order to fuse a sequence of representations generated by Transformers~\cite{vaswani2017attention}, there are two major types of cross-modal fusion modules: \textbf{merged fusion}, and \textbf{cross fusion}. In merged fusion, the two modalities are concatenated and then fed into a Transformer block \cite{li2020unicoder,li2019visualbert,Luo2020UniVL,wang2022git}. 
In cross fusion, the two modalities are fed into different Transformer blocks separately, where cross attention Transformers are used to allow cross-modal interaction \cite{lu2019vilbert,seo2022end,seo2021look,tan2019lxmert,tsai2019multimodal}. Besides, some recent works propose variants of cross-modal fusion modules that use bottleneck tokens \cite{nagrani2021attention} or prune single-modal units \cite{wang2022multimodal} to control the flow of cross-modal interaction.

\section{Audio-Visual Video Captioning}
In this section, we propose our methods for audio-visual video captioning. In Sec.~\ref{subsec:overview},
we present an overview of our framework. In Sec.~\ref{subsec:multimodal-pre-train}, we describe our MBP loss to balance different modalities during pre-training.
In Sec.~\ref{subsec:cross-modal-fusion}, we investigate different cross-modal fusion modules and propose a local-global fusion module to improve information flow between audio and video at different granularities.

\subsection{Framework Overview}
\label{subsec:overview}

As shown in Fig.~\ref{fig:framework}, we follow the common practice in video captioning \cite{Luo2020UniVL,seo2022end} and use an encoder-decoder framework including four main modules: a video encoder, an audio encoder, a cross-modal encoder, and a caption decoder, all of which are Transformer architectures \cite{vaswani2017attention}. 

Given a video, the video encoder extracts a sequence of video embeddings ${\varphi}^v \in \mathbb{R}^{N_v \times D}$ from the raw image frames, and the audio encoder extracts a sequence of audio embeddings ${\varphi}^a \in \mathbb{R}^{N_a \times D}$ from the associated audio spectrogram, where $N_v$ and $N_a$ are the numbers of video tokens and audio tokens, and $D$ is feature dimension. Then, we employ a cross-modal encoder $f_{\Theta}(\cdot)$ to generate multi-modal embeddings ${\varphi}^c \in \mathbb{R}^{(N_a+N_v) \times D}$  for cross-modal interaction. Finally, we use a decoder $g_{\Theta'}(\cdot)$ conditioned on  ${\varphi}^c$ to output the captions auto-regressively. By default, we use Video Swin Transformer \cite{liu2022video} as the video encoder, and Audio Spectrogram Transformer \cite{gong2021audio} as the audio encoder.

Inspired by \cite{seo2022end}, we design two tasks for caption generation during pre-training:  Predict Current Caption (PCC) and Predict Next Caption (PNC). Given the audio-visual embeddings as the context of the caption decoder, we feed two distinct \texttt{BOS} (Beginning of Sentence)  tokens to initiate caption generation, namely \texttt{BOS1} and \texttt{BOS2}, {which initiate the prediction of the current and next caption respectively.} 
{The PNC task enforces the model to} anticipate future events,  
which is more challenging and requires higher level of semantic understanding. Note that this multi-task training is only used for pre-training. For downstream fine-tuning, we only feed \texttt{BOS1} token to the decoder as the goal is to predict the current caption.

\subsection{{Modality Balanced Pre-training}}
\label{subsec:multimodal-pre-train}

With captions as supervision, a commonly used objective is to minimize the negative log-likelihood of generating the ground-truth caption:
\begin{equation}
    L = \mathcal{L}(g_{\Theta'}(f_{\Theta}({\varphi}^a, {\varphi}^v)), y),
\label{eq:naive-pre-train}
\end{equation}
where $\mathcal{L}$ is the cross entropy loss and $y$ is the ground-truth caption. We refer to this loss as audio-video decoder loss, as both audio and video features are input into the decoder.

Prior works \cite{akbari2021vatt,miech2020end,seo2022end,wang2022git,zellers2022merlot,zellers2021merlot} have proved that large-scale pre-training is essential to multi-modal learning. Thus, we pre-train our audio-visual video captioning model on a large-scale video dataset before fine-tuning it on the downstream captioning datasets. 
Although there are some large-scale video datasets for video-language pre-training 
(\eg, HowTo100M \cite{miech2019howto100m} and HD-VILA-100M \cite{xue2022advancing}), they do not contain manually annotated captions and use ASR transcripts as text supervision.
One significant issue based on our experiments (see Fig.~\ref{fig:training-loss}) is that the model tends to learn only from the audio and ignore the video modality, since the ASR caption can be solely derived from human speech in the audio. Although we include two tasks for current and future caption predictions, the model still favors the audio features when predicting the current caption.

To address this issue, we propose the MBP loss to balance different modalities for multi-modal training. In essence, we add losses for audio-only and video-only predictions 
to help improve the representation learning of the two modalities.
To measure how well the model exploits a certain modality, we first define the following two losses, \ie, audio-only decoder loss and video-only  decoder loss:
\vspace{-1mm}
\begin{equation}
\begin{aligned}
     L_a &= \mathcal{L}(g_{\Theta'}(f_{\Theta}( {\varphi}^a, \mathbf{0})), y), \\
     L_v &= \mathcal{L}(g_{\Theta'}(f_{\Theta}(\mathbf{0}, {\varphi}^v)), y),
\end{aligned}
\end{equation}
where we set video embeddings to all-zeros to get the audio-only decoder loss $L_a$, and set audio embeddings to all-zeros to get the video-only decoder loss $L_v$ based on {Eq.} \eqref{eq:naive-pre-train}. We refer to $L_a$ or $L_v$ as mono-modal losses. If a mono-modal loss is small, it means that the corresponding modality is well utilized by the model. 
We then measure the gap between the multi-modal loss and the mono-modal losses by {a Mono-to-Multi Discrepancy (MMD) index}:
\begin{equation}
    G_a = (L_a - L)^2; G_v = (L_v - L)^2,
    \label{eq:gap}
\end{equation}
where $G_a$ and $G_v$ measure the discrepancy between audio-only/video-only decoder loss and audio-video decoder loss, respectively. 
Inspired by G-Blend \cite{wang2020makes} that uses Overfitting-to-Generalization Ratio (OGR) to iteratively update training weights for different modalities, we guide the weights of mono-modal losses based on whether the modality is well utilized by the model:
\begin{equation}
    L_{pretrain} = L + w_a L_a + w_v L_v,
    \label{eq:weighted_pre-train}
\end{equation}
where $w_a$ and $w_v$ are weights updated over iterations:
\begin{equation}
    w_m^{(t)} = \beta w_m^{(t-1)} + (1-\beta) \Tilde{w}_m^{(t)}, m \in \{a, v\},
    \label{eq:weight_update}
\end{equation}
where $\beta \in (0,1)$ is a smoothing hyperparameter, $t$ is the iteration number, and $\Tilde{w}_m^{(t)}$ is obtained using a softmax function over the gap between each mono-modal loss and the multi-modal loss of the current iteration:
\begin{equation}
     \Tilde{w}_m^{(t)} = \frac{\exp{(\alpha  G_m^{(t)})}}{\sum_{m'}\exp{(\alpha G_{m'}^{(t)})}}, m \in \{a, v\},
     \label{eq:weight_softmax}
\end{equation}
where $\alpha>0$ is a temperature hyperparameter. If $G_m$ is large for a certain modality on the pre-training dataset, we will assign a higher weight $w_m$ for modality $m$ in {Eq.} \eqref{eq:weighted_pre-train}. By doing this, we enforce the model to give more attention to modality $m$ and mitigate its overspecialization to the other modality. When the optimization progresses and the gaps $G_a$ and $G_v$ change over time, we dynamically adjust the coefficients $w_a$ and $w_v$ according to {Eq.} \eqref{eq:weight_update} to 
increase the weight of the underfitted modality. Our strategy can be easily extended to more than two modalities, which is out of the scope of this work. Empirically, we found the MBP loss not sensitive to the two hyperparameters $\beta$ and $\alpha$. We set $\beta=0.99$ and $\alpha=10$ for all the experiments.

\begin{figure*}[tbh]
     \centering
     \begin{subfigure}[b]{0.25\textwidth}
         \centering
         \includegraphics[width=\textwidth,height=38mm]{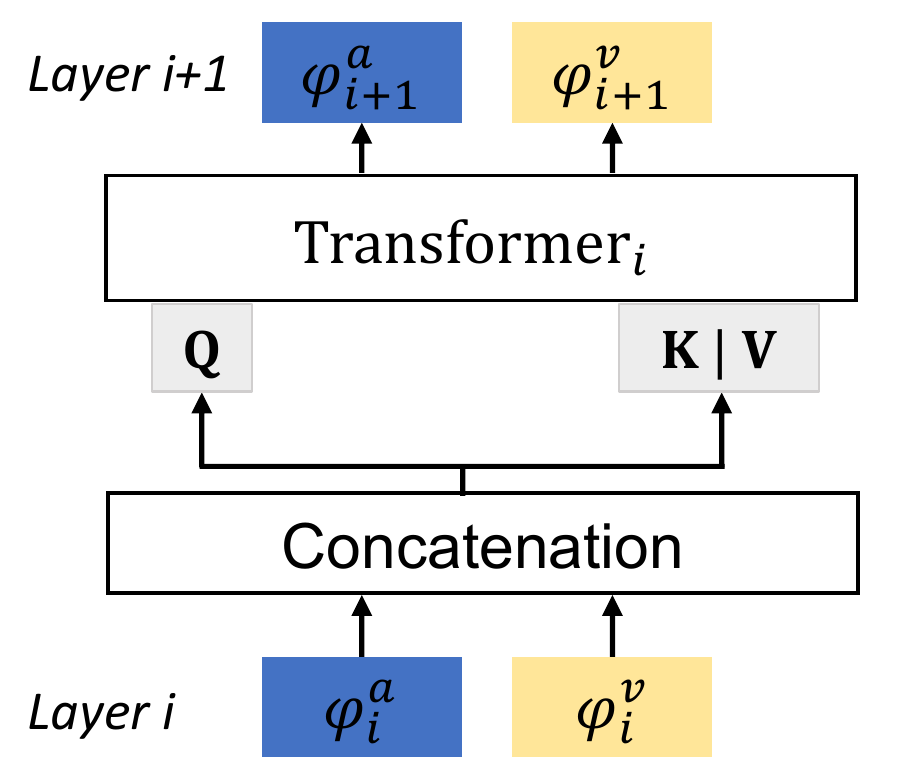}
         \caption{Merged Fusion}
         \label{subfig:merged-attention}
     \end{subfigure}
     \hfill
     \begin{subfigure}[b]{0.35\textwidth}
         \centering
         \includegraphics[width=\textwidth,height=38mm]{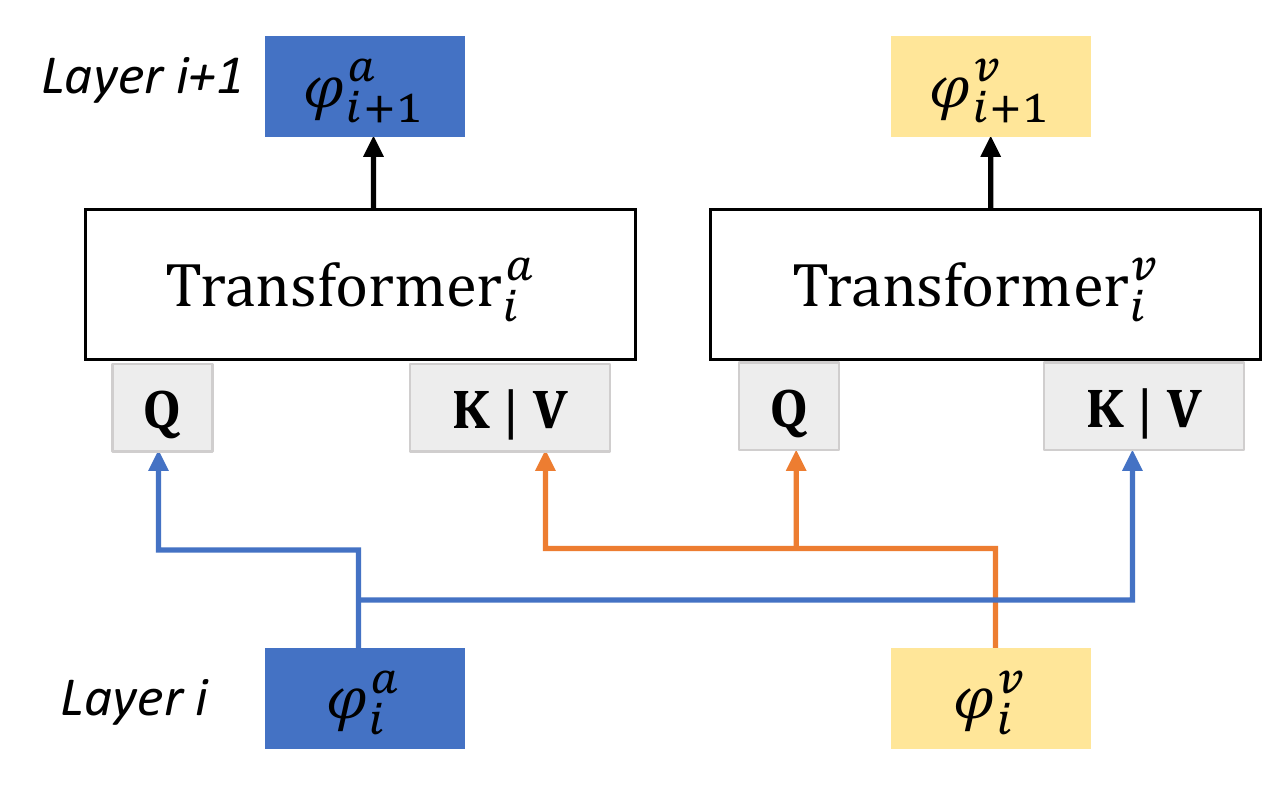}
         \caption{Cross Fusion}
         \label{subfig:cross-attention}
     \end{subfigure}
     \hfill
     \begin{subfigure}[b]{0.37\textwidth}
         \centering
         \includegraphics[width=\textwidth,height=38mm]{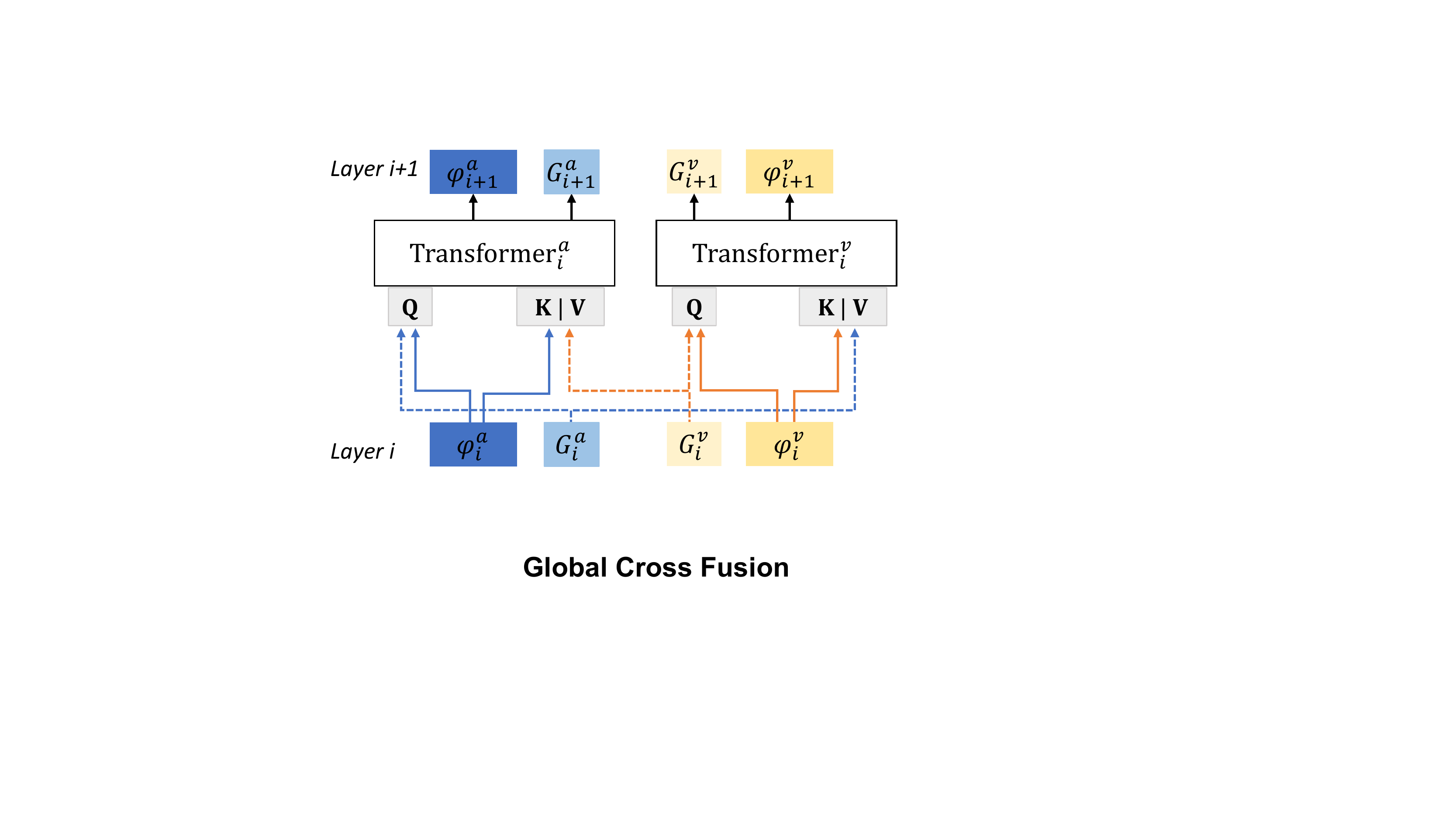}
         \caption{Global Cross Fusion}
         \label{subfig:global-attention}
     \end{subfigure}
        \caption{Three different cross-modal fusion designs. Superscripts $a, v$ denote audio and video modalities. $\Q$, $\K$, $\V$ represent query, key, and value in multi-head attention models. For clarity, we use dot lines to denote the flow of global tokens in (c). 
        }
        \label{fig:cross-modal-fusion}
\end{figure*}

\subsection{{Cross-Modal Fusion: Beyond Cross Attention}}
\label{subsec:cross-modal-fusion}
In addition to the MBP objective, the cross-modal encoder is another component where different modalities interact with each other. In this section, we explore different design choices for the cross-modal fusion module to better leverage the audio modality for video captioning.

\vspace{-3mm}
\subsubsection{Background}
\label{subsubsec:background}
We first briefly introduce some relevant concepts of cross-modal fusion modules.
Given $N_q$ query vectors of dimension $d$, $\Q\in \Re^{N_q \times d}$, and $N_v$ key-value pairs, $\K \in \Re^{N_v \times d}$ and $\V \in \Re^{N_v \times d}$, an attention function  maps queries to output vectors with a scaled dot product:
\vspace{-2mm}
\begin{equation}
\vspace{-2mm}
    \Att(\Q, \K, \V) = \Softmax(\frac{\Q\K^T}{\sqrt{d}})\V.
\end{equation}
\textbf{Multi-Head Attention} (MHA) is a module with multiple heads of attention functions, and a \textbf{Transformer layer} consists of an MHA and a Feed-Forward Block (FFB). The keys and values are usually the same features in the Transformer layer. For simplicity, we denote a Transformer layer by:
\vspace{-1mm}
\begin{equation}
    \Transformer(\X, \Y) = \FFB(\MHA(\X,\Y,\Y)).
    \vspace{-2mm}
\end{equation}

\subsubsection{Cross-Modal Fusion}
\label{subsubsec:cross-modal-fusion}
 
As mentioned in Sec.~\ref{subsec:overview}, we get a sequence of audio tokens ${\varphi}^a \in \mathbb{R}^{N_a \times D}$ from the audio encoder, and a sequence of video tokens ${\varphi}^v \in \mathbb{R}^{N_v \times D}$ from the video encoder. Then we feed audio and video tokens into a cross-modal encoder for cross-modal fusion. 
Below we introduce two popular fusion methods: merged fusion and cross fusion. 

In merged fusion (Fig.~\ref{subfig:merged-attention}), the tokens of two modalities are concatenated and then passed into Transformer blocks:
\vspace{-1mm}
\begin{equation}
   \varphi^a_{i+1}, \varphi^v_{i+1} =  \Transformer_i([\varphi^a_i ;\varphi^v_i], [\varphi^a_i ;\varphi^v_i]).
   \label{eq:self-attention}
\end{equation}
Audio tokens $\varphi^a_i$ and video tokens $\varphi^v_i$ are the inputs of the $i$-th Transformer layer. 
Each audio token can attend to all audio and video tokens, and it is the same for video tokens. 

In cross fusion (Fig.~\ref{subfig:cross-attention}), each modality has its own Transformer layers, and different modalities exchange information via cross attention, \ie, one modality is used as the context (keys and values) of the other modality: 
\begin{equation}
    \begin{aligned}
        \varphi^a_{i+1} =  \Transformer_i^a(\varphi^a_i, \varphi^v_i), \\
        \varphi^v_{i+1} =  \Transformer_i^v(\varphi^v_i, \varphi^a_i).
    \end{aligned}
    \label{eq:cross-attention}
\end{equation}
Each layer has two modality-specific Transformers.

\vspace{-1mm}
\subsubsection{Global Cross Fusion}
\label{subsubsec:global-cross-att}

We use the term ``local fusion" to denote the two fusion modules introduced in Sec.~\ref{subsubsec:cross-modal-fusion} as the interaction is among local tokens, regardless of intra-modality or inter-modality. However, only using local interaction can be sub-optimal, as local tokens are often noisy and may be less informative for cross-modal interaction. 
To select more salient features for cross-modal fusion, we introduce global tokens $G^a$ and $G^v$ for each modality as shown in Fig.~\ref{subfig:global-attention}. Global tokens serve as a global representation of the audio clip or the video clip. 
In global cross fusion, instead of using all video tokens as the context for audio tokens, 
we use the global video token as the context:
\begin{equation}
    \begin{aligned}
        \varphi^a_{i+1}, G^a_{i+1} =  \Transformer_i^a([\varphi^a_i ; G^a_i], [\varphi^a_i ; G^v_i]), \\
        \varphi^v_{i+1}, G^v_{i+1} =  \Transformer_i^v([\varphi^v_i ; G^v_i], [\varphi^v_i ; G^a_i]).
    \end{aligned}
    \label{eq:global}
\end{equation}

Here, we restrict all cross-modal attention flow to be via these global tokens, and local tokens will only be used for intra-modal attention.
The global token of the first cross layer is learnable and is initialized with a Gaussian distribution, the same as the \texttt{CLS} token in BERT \cite{kenton2019bert}. 

\subsubsection{Local-Global Fusion}
Local and global tokens capture information in different granularities. Local audio and video tokens capture local features such as words in the speech or objects in a video frame, while global audio and video tokens capture high-level concepts like sounds of laughter or people gathering on a street. 
To take advantage of both local and global information, we propose to combine local fusion and global cross fusion. Let $\varphi^{a(G)}_{i+1}$ and $\varphi^{v(G)}_{i+1}$ denote the embeddings from global cross fusion in {Eq.} \eqref{eq:global}, and $\varphi^{a(L)}_{i+1}$ and $\varphi^{v(L)}_{i+1}$ denote the embeddings from local fusion in {Eq.} \eqref{eq:self-attention} or {Eq.} \eqref{eq:cross-attention}, we compute an average of these two embeddings before feeding them into the next fusion layer, \ie,
\begin{equation}
    \varphi^m_{i+1} = (\varphi^{m(G)}_{i+1} + \varphi^{m(L)}_{i+1})/2, \; m \in \{a, v\}.
\end{equation}
This unified fusion module is named as ``local-global fusion", and is able to progressively refine the tokens using both local and global guidance. 
We name this variant with merged fusion as ``local-global merged fusion", and the variant with cross fusion as ``local-global cross fusion".

\section{Experiments}
\subsection{Pre-training and Downstream Datasets}
We use the HowTo100M dataset \cite{miech2019howto100m} for pre-training, and evaluate the proposed method on four video captioning datasets, including YouCook2 \cite{zhou2018towards}, MSRVTT \cite{xu2016msr}, VATEX \cite{wang2019vatex} and ActivityNet-Captions \cite{shi2019dense}. We use four metrics, \ie, BLEU-4 (B) \cite{papineni2002bleu}, METEOR (M) \cite{banerjee2005meteor}, ROUGE-L (R) \cite{lin2004rouge}, and CIDEr (C) \cite{vedantam2015cider}, for evaluation.

\textbf{HowTo100M} \cite{miech2019howto100m} consists of $1.2M$ instructional videos from YouTube. We download videos with ASR transcripts and audio from YouTube, and remove currently unavailable videos, resulting in $1.08M$ videos in total. Following \cite{Luo2020UniVL}, we start with a single ASR sentence and iteratively expand the length of video clip forward or backward  by adding next or previous sentences until the clip is longer than 5 seconds.

\textbf{YouCook2} \cite{zhou2018towards} contains $2,000$ cooking videos with $15.4k$ video clips. Each video clip is annotated with a single sentence.
\textbf{MSRVTT} \cite{xu2016msr} contains $10k$ open-domain video clips for video captioning. Each video clip is annotated with $20$ captions. 
\textbf{VATEX} \cite{wang2019vatex} consists of $41,250$ videos. Each video is annotated with $10$ English captions and $10$ Chinese captions, and we use the English captions. \textbf{ActivityNet-Captions} \cite{krishna2017dense} is a large-scale video paragraph captioning dataset consisting of $100k$ temporally localized captions for $20k$ long videos. To be consistent with the other datasets, we train our model on sentence-level captions. Following \cite{zhou2018end}, we compose paragraph-level captions by simply concatenating sentence-level captions and evaluate the captioning performance at paragraph-level in testing. Please refer to the appendix for more details. 

\spar{Audio Relevance Analysis.} 
The performance of video captioning on downstream datasets is also related to the language styles of the annotations, so we design two metrics, Speech Coverage Rate (SCR) and Audio Relevance Score (ARS), to analyze how related the annotated captions are to audio. Concisely, SCR measures the percentage of words in ground-truth captions that also occur in speech transcripts; ARS measures the relevance of the captions to audio. ARS is empirically calculated based on the word frequencies on audio captioning \cite{drossos2020clotho,kim2019audiocaps} and image captioning \cite{chen2015microsoft} datasets. If the frequency of a word is much higher in audio captions than that in image captions, we assume that this word is highly related to audio modality and assign a high score to a sentence with the word. We include more details in the appendix.  Examples of the words with a high ARS are \textit{``loudly", ``snore", ``footstep"}, \etc. Those words are usually related to acoustic events, scenes, or sound patterns, and require more holistic or long-term information from the audio for the model to understand the context. Tab.~\ref{tab:language-stats} lists the SCR and ARS on downstream datasets. We notice that a large portion of the captions on YouCook2 are mentioned in speech. Conversely, though only a small amount of captions are covered by speech, the captions on VATEX are most relevant to audio modality. 

\begin{table}[!t]
\centering
\resizebox{0.85\linewidth}{!}{
\begin{tabular}{c|cccc}
\toprule
         & YouCook2 & MSRVTT & VATEX & ActivityNet \\ 
\midrule
SCA  & \textbf{48.67}    & 14.7   & 5.53  &   12.63          \\  
ARS      & 1.202    & 2.903  & \textbf{3.692} & 3.487       \\ 
\bottomrule
\end{tabular}}
\vspace{-2mm}
\caption{\small{The Speech Coverage Rate (SCA, \%) and Audio Relevance Score (ARS) on the downstream datasets.}}
\vspace{-4mm}
\label{tab:language-stats}
\end{table}

\begin{table*}[tbh]
    \centering
    \resizebox{0.9\linewidth}{!}{
\begin{tabular}{ccc|cccc|cccc|cccc|cccc}
\toprule
\multirow{2}{*}{MBP} & \multirow{2}{*}{PCC} & \multirow{2}{*}{PNC} & \multicolumn{4}{c|}{YouCook2}   & \multicolumn{4}{c|}{MSRVTT}    & \multicolumn{4}{c|}{VATEX}     & \multicolumn{4}{c}{ActivityNet-Captions} \\
                                    &                      &                      & B     & M     & R     & C      & B     & M     & R     & C     & B     & M     & R     & C     & B      & M      & R     & C     \\
\midrule
G-Blend \cite{wang2020makes}  &  \cmark                                  & \cmark                        & 19.4                  & 23.4                  & 48.6                  & 208.5                 & 44.0                  & 29.6                  & 62.4                  & 55.1      & 38.0                  & 24.9                  & 52.1                  & 68.7                  & 11.6                  & 16.1                  & 30.6                  & 25.3            \\
\midrule
                                & \cmark                     &                   & 15.9          & 20.2          & 43.0          & 166.8          & 41.4          & 28.5          & 60.7          & 47.6          & 31.2          & 21.9          & 47.7          & 50.7          & 10.0          & 14.6          & 28.4          & 20.1 \\
\cmark                                  & \cmark                    &                     & 18.5          & 22.7          & 47.0          & 192.4          & 44.7          & 29.9          & 62.7          & 53.5          & 36.9          & 24.7          & 51.7          & 67.5          & 11.3          & 15.5          & 30.2          & 24.7   \\
                               & \cmark                    & \cmark                    & 17.2                  & 21.7                  & 45.8                  & 184.2                 & 42.0                  & 28.4                  & 60.8                  & 48.4                  & 31.2                  & 22.0                  & 48.0                  & 51.4                  & 10.4                  & 14.9                  & 28.9                  & 20.2    \\
\cmark                                  & \cmark                    & \cmark                    & \textbf{20.6} & \textbf{24.2} & \textbf{49.6} & \textbf{217.0} & \textbf{46.0} & \textbf{30.6} & \textbf{64.0} & \textbf{57.0} & \textbf{38.8} & \textbf{25.9} & \textbf{52.9} & \textbf{73.5} & \textbf{11.7} & \textbf{16.1} & \textbf{30.7} & \textbf{26.1}   \\
\bottomrule
\end{tabular}}
\vspace{-2mm}
\caption{
Ablation studies on multi-modal pre-training {with our audio-visual captioning framework from Fig.~\ref{fig:framework}}. MBP: Modality Balanced Pre-training; PCC: Predict Current Caption; PNC: Predict Next Caption.}
\label{tab:multimodal-pre-train}
\end{table*}

\begin{table}[!t]
\vspace{-4mm}
    \centering
    \resizebox{\linewidth}{!}{
\begin{tabular}{c|cccc|cccc}
\toprule
\multirow{2}{*}{Fusion} & \multicolumn{4}{c|}{YouCook2}                                  & \multicolumn{4}{c}{VATEX}                                    \\
                        & B             & M             & R             & C              & B             & M             & R             & C             \\ \midrule
cross                   & 19.5          & 23.4          & 48.9          & 211.2          & 37.8          & 24.7          & 52.0          & 68.9          \\
merged                  & 19.9          & 23.6          & 49.1          & 210.7          & 38.4          & 25.1          & 52.3          & 71.1          \\
global                  & 18.6          & 22.9          & 48.0          & 202.3          & \textbf{39.2} & 25.5          & 52.8          & 72.6          \\
local-global cross      & 19.9          & 23.9          & 49.2         & 213.9         & 38.6          & 25.8          & \textbf{53.0}          & 73.4          \\
local-global merged     & \textbf{20.6} & \textbf{24.2} & \textbf{49.6} & \textbf{217.0} & 38.8          & \textbf{25.9} & {52.9} & \textbf{73.5} \\ \bottomrule
\end{tabular}}
\vspace{-2mm}
\caption{{Ablation studies on cross-modal fusion modules.}}
\vspace{-2mm}
\label{tab:crossmodal-fusion}
\end{table}

\subsection{Experimental Setup}

\npar{Video Encoder}: We sample 16 frames from each video clip. 
The frames are resized and cropped into images of size $224 \times 224$. 
Each video clip with the size of $16\times 224 \times 224 \times 3$ is fed into the Video Swin Transformer \cite{liu2022video} initialized with the weights pre-trained on Kinetics 600 \cite{carreira2018short} and tokenized into $N_v = 8\times 7 \times 7 = 392$ video tokens. Then we add a linear layer to project the dimension of each video token to $D=768$, to be consistent with the other modules. 

\npar{Audio Encoder}: We first extract log mel spectrogram of the raw audio. Following \cite{zellers2022merlot}, each audio is resampled to 22,050Hz and divided into frames of 1536 samples with hop length of $588$. Then we apply 64 mel-scale filters. 
We use a 12-layer Transformer on audio spectrogram to output $N_a = 64$ audio tokens with feature dimension of 768. 

\npar{Cross Encoder}: We use a 3-layer Transformer as the cross-modal encoder. In our experiments, we comprehensively compare the results of the cross-modal fusion methods introduced in Sec.~\ref{subsec:cross-modal-fusion}.
Overall, merged fusion outperforms cross fusion, and local-global merged fusion achieves the best performance. 
For other experiments, we use local-global merged fusion unless explicitly stated otherwise. 

\npar{Caption Decoder}: We use a 3-layer Transformer as the caption decoder. In training, we use causal masking to ensure that only history inputs are used. In testing, we use beam-search with beam width of 5 for caption generation.

\npar{Training Details:} We pre-train the model on HowTo100M for 100 epochs using Adam optimizer \cite{Kingma:ICLR15}. For each video, we sample three video-caption pairs from a long video in one iteration. We pre-train the model on 64 Nvidia Tesla A100 GPUs and it takes around five days. The base learning rate is $10^{-4}$ and we use a linear decay learning rate schedule with a warm-up of $10\%$ training epochs as in \cite{Luo2020UniVL}. 
For fine-tuning, we set the initial learning rate as $10^{-5}$. We train the model for 5 epochs on YouCook2 and MSRVTT, 10 epochs on VATEX, and 30 epochs on ActivityNet-Captions. It is noted that we employ modality balancing during pre-training but not during downstream fine-tuning, to allow the model to adapt to the more informative modality for caption generation. We have included more details on the model architecture and implementation in the appendix.

\begin{figure}[!t]
    \centering
    \includegraphics[width=0.95\linewidth]{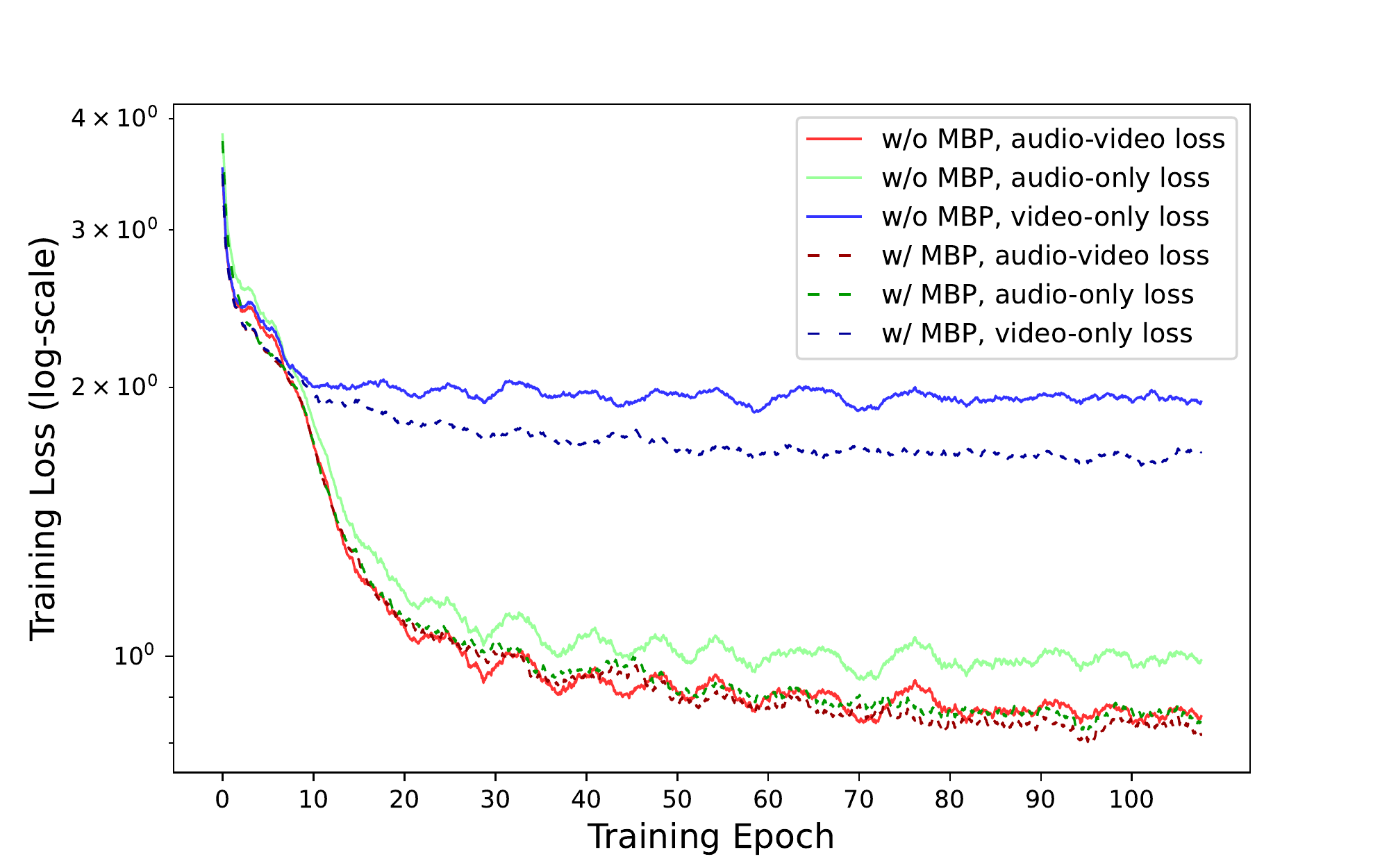}
    \vspace{-2mm}
    \caption {\small{The pre-training losses without and with MBP. (Solid lines: losses without MBP. Dotted lines: losses with MBP.)}} \vspace{-2mm}
    \label{fig:training-loss}
    \vspace{-2mm}
\end{figure}

\subsection{Results and Discussions}

\npar{Multi-modal pre-training objectives.} 
We evaluate the effects of our pre-training objectives in Tab.~\ref{tab:multimodal-pre-train}. Our proposed MBP improves the performance by a large margin on four datasets. The addition of Predicting Next Caption leads to a remarkable boost as well. 
{We also compare MBP with G-Blend \cite{wang2020makes}: G-Blend aims to reduce overfitting, which happens when the model performs well on the training set but fails to generalize to new data, whereas MBP aims to reduce overspecialization, which happens when the model performs well on a single modality but fails to learn from the other modalities. Hence, G-Blend updates mono-modal weights by computing Overfitting-to-Generation Ratio while MBP uses Mono-to-Multi Discrepancy. According to the results in Tab.~\ref{tab:multimodal-pre-train}, G-Blend also improves the performance, but MBP consistently outperforms G-Blend, showing that MBP is more suitable to avoid the issue of overspecialization in audio-visual captioning pre-training.}
In Fig.~\ref{fig:training-loss}, we show the trend of training losses with and without MBP, including multi-modal loss (audio-video loss) and mono-modal losses (audio-only loss and video-only loss).  Note that for the experiments without MBP, audio-only loss and video-only loss are only computed for analysis and not back-propagated. Although the two audio-video losses have similar scales, both audio-only loss and video-only loss are reduced
with MBP, especially for video-only loss. We conjecture that the model reduces overspecialization to the audio modality and learns to better utilize the video modality with MBP. Though there is only a small decrease in the audio-video training loss, the performance on downstream tasks are significantly improved (see Tab.~\ref{tab:multimodal-pre-train}), which demonstrates the effectiveness of our pre-training objective.

\begin{table*}[tbh]
    \centering
    \resizebox{0.9\linewidth}{!}{
\begin{tabular}{c|cccc|cccc|cccc|cccc}
\toprule
\multirow{2}{*}{Inputs} &  \multicolumn{4}{c|}{YouCook2} & \multicolumn{4}{c|}{MSRVTT} & \multicolumn{4}{c|}{VATEX} & \multicolumn{4}{c}{ActivityNet} \\
                &  B     & M     & R    & C     & B     & M    & R    & C    & B    & M    & R    & C    & B      & M      & R     & C     \\
\midrule
A    &        13.4          & 17.6          & 38.5          & 138.9          & 32.3          & 23.9          & 54.3          & 24.8          & 15.4          & 15.6          & 37.6          & 14.2          & 6.3           & 11.2          & 24.5          & 6.6           \\
V            & 11.7          & 18.0          & 41.5          & 139.2          & 42.6          & 28.5          & 61.0          & 52.0          & 36.4          & 24.6          & 51.2          & 67.3          & 9.7           & 14.4          & 27.8          & 21.3          \\
\hline
V+A             & \textit{20.6} & \textit{24.2} & \textit{49.6} & \textit{217.0} \small{(77.8$\uparrow$)}  & \textit{46.0} & \textbf{30.6} & \textit{64.0} & \textit{57.0} \small{(5.0$\uparrow$)}  & \textit{38.8} & \textit{25.9} & \textit{52.9} & \textit{73.5} \small{(6.2$\uparrow$)}  & \textit{11.7} & \textbf{16.1} & \textit{30.7} & \textit{26.1} \small{(4.8$\uparrow$)} \\
V+T             & 19.1               & 23.3               & 48.7            & 205.6  \small{(66.4$\uparrow$)}               & 43.3               & 29.1                & 61.7               & 53.4       \small{(1.4$\uparrow$)}            & 37.7                 & 25.2                & 52.0                & 69.0    \small{(1.7$\uparrow$)}              & 11.2                 & \textit{15.9}                 & 29.4                & 24.9       \small{(3.6$\uparrow$)}           \\ 
V+A+T            & \textbf{20.9} & \textbf{24.4} & \textbf{49.9} & \textbf{221.6} \small{(82.4$\uparrow$)}  & \textbf{46.4} & \textit{30.2} & \textbf{64.1} & \textbf{57.3}  \small{(5.3$\uparrow$)}  &   \textbf{39.1}	  & \textbf{26.3}	& \textbf{53.4}  & 	\textbf{73.7} \small{(6.4$\uparrow$)}                                                                            & \textbf{11.8} & \textbf{16.1} & \textbf{30.9} & \textbf{26.4}  \small{(5.1$\uparrow$)} 
\\
\bottomrule 
\end{tabular}}
\vspace{-2mm}
\caption{The performance when we input different modalities (V: video, A: audio, T: text). We show the improvement of multi-modal video captioning over the video-only method in terms of CIDEr in the parenthesis.}
\label{tab:video-audio-text}
\end{table*}

\begin{table*}[tbh]
\begin{subtable}{0.48\textwidth}
\centering
\resizebox{0.99\linewidth}{!}{
\begin{tabular}{ccccccc}
\toprule
Method    & Pre-training dataset         & Inputs & B    & M    & R    & C   \\
\midrule
GIT2  \cite{wang2022git} $^{\dagger}$    & 12.9B image-text pairs                 & V      & 9.4  & 15.6 & 37.5 & 131.2 \\
\hline
MART \cite{pei2019memory}     & -                           & V      & 8.0  & 15.9 & -    & 36.0  \\
SwinBert \cite{lin2022swinbert} & -                          & V      & 9.0  & 15.6 & 37.3 & 109.0 \\
VideoBert \cite{sun2019videobert} & YouTube                     & V      & 4.0  & 11.0 & 27.5 & 49.0  \\
M-MASS \cite{huang2020multimodal}   & YT8M-cook+Recipe1M          & V+T    & 12.0 & 18.3 & 39.0 & 123.0 \\
ActBert \cite{zhu2020actbert}  & HT100M                      & V      & 5.4  & 13.3 & 30.6 & 65.0  \\
Value  \cite{li2021value}   & HT100M+TV & V+T    & 12.4 & 18.8 & 40.4 & 130.4 \\
UniVL \cite{Luo2020UniVL}    & HT100M                      & V+T    & {17.4} & {22.4} & 46.5 & 181.0 \\
MV-GPT \cite{seo2022end}   & HT100M                       & V+T    & 21.9$^{\ast}$ & 27.1$^{\ast}$  & {49.4} & \textit{221.0} \\
\midrule
Ours    & HT100M                       & V+A    & \textit{20.6} & \textit{24.2} & \textit{49.6} & {217.0} \\
Ours    & HT100M                       & V+T    & 19.1               & 23.3               & 48.7            & 205.6 \\
Ours    & HT100M                       & V+A+T &    \textbf{20.9} & \textbf{24.4} & \textbf{49.9} & \textbf{221.6} \\
\bottomrule
\end{tabular}
}
\label{subtab:sota-yc2}
\vspace{-2mm}
\caption{YouCook2 }
\end{subtable}
\hfill
\begin{subtable}{0.48\textwidth}
\centering
\resizebox{0.99\linewidth}{!}{
\begin{tabular}{ccccccc}
\toprule
Method    & Pre-training dataset & Inputs & B    & M    & R    & C   \\
\midrule
GIT2  \cite{wang2022git} $^{\dagger}$   & 12.9B image-text pairs          & V      & 54.8 & 33.1 & 68.2 & 75.9  \\
\hline
Open-Book \cite{zhang2021open} & -                   & V      & 33.9 & 23.7 & 50.2 & 52.9  \\
SwinBert \cite{lin2022swinbert} & -                   & V      &  {45.4} & \textbf{30.6} & \textit{64.1} & 55.9  \\
MMCNN    \cite{tian2019audio}                                           & -                     & V+A    & 42.7 & 28.5 & 61.5 & 47.2  \\
MGSA    \cite{chen2019motion}                                           & -                     & V+A    & {45.4} & 28.6 & - & 50.1  \\
Decembert \cite{tang2021decembert} & HT100M              & V+T    & 45.2 & {29.7} & \textbf{64.7} & 52.0  \\
UniVL \cite{Luo2020UniVL}    & HT100M              & V+T    & 41.8 & 28.9 & 60.8 & 50.0  \\
MV-GPT  \cite{seo2022end}  & HT100M              & V+T    & 48.9$^{\ast}$ & 38.7$^{\ast}$ & 64.0 & \textbf{60.0}  \\
\midrule
Ours      & HT100M              & V+A    & \textit{46.0} & \textbf{30.6} & 64.0 & {57.0} \\
Ours      & HT100M              & V+T &43.3               & 29.1                & 61.7               & 53.4      \\
Ours      & HT100M              & V+A+T    &  \textbf{46.4} & \textit{30.2} & \textit{64.1} & \textit{57.3} \\
\bottomrule
\end{tabular}
}
\label{subtab:sota-msrvtt}
\caption{MSRVTT}
\end{subtable}
\begin{subtable}{0.48\textwidth}
\centering
\resizebox{0.99\linewidth}{!}{
\begin{tabular}{ccccccc}
\toprule
Method    & Pre-training dataset & Inputs & B    & M    & R    & C   \\
\midrule
GIT2  \cite{wang2022git} $^{\dagger}$   & 12.9B image-text pairs          & V      & 42.7 & 28.8 & 56.5 & 94.5  \\
\hline
SwinBert \cite{lin2022swinbert} & -                   & V      & {38.7} & \textit{26.2} & \textit{53.2} & {73.0} \\
MGRMP \cite{chen2021motion}                                           & -                     & V   & 34.2 & 23.5 & 50.3 & 57.6  \\
Value  \cite{li2021value}   & HT100M+TV & V+T    & 32.9 & 24.1 & 50.1 & 58.1 \\
\midrule
Ours      & HT100M              & V+A    & \textit{38.8} & {25.9} & {52.9} & \textit{73.5} \\ 
Ours      & HT100M              & V+T  & 37.7                 & 25.2                & 52.0                & 69.0 \\
Ours      & HT100M              & V+A+T    & \textbf{39.1}	  & \textbf{26.3}	& \textbf{53.4}  & 	\textbf{73.7} \\
\bottomrule
\end{tabular}
}
\label{subtab:sota-vatex}
\caption{VATEX}
\end{subtable}
\hfill
\begin{subtable}{0.48\textwidth}
\centering
\resizebox{0.99\linewidth}{!}{
\begin{tabular}{ccccccc}
\toprule
Method         & Pre-training dataset & Inputs & B              & M              & R              & C              \\ \midrule
VTransformer \cite{zhou2018end}   & ActivityNet          & V      & 9.31           & 15.54          & -              & 21.33          \\
Transformer-XL \cite{dai2019transformer} & ActivityNet          & V      & 10.25          & 14.91          & -              & 21.71          \\
MART    \cite{pei2019memory}  $^{\ddagger}$     & ActivityNet          & V      & 9.78           & 15.57          & 30.63             & 22.16          \\
COOT    \cite{ging2020coot}    $^{\ddagger}$     & HT100M               & V      & {10.85}          & \textit{15.99}          & \textbf{31.45} & \textbf{28.19} \\ \midrule
Ours            & HT100M               & V+A    & \textit{11.73} & \textbf{16.14} & {30.68}          & {26.11}          \\ 
Ours            & HT100M               & V+T    & 11.22                 & 15.94               & 29.40                & 24.92     \\
Ours            & HT100M               & V+A+T    & \textbf{11.83} & \textbf{16.14} & \textit{30.93} & \textit{26.38} \\
\bottomrule
\end{tabular}
}
\label{subtab:sota-activitynet}
\caption{ActivityNet-Captions}
\end{subtable}
\vspace{-2mm}
\caption{\small{Comparison to SOTA. The top two results are in \textbf{bold} and \textit{italic}. $\dagger$ use a large-scale annotated image captioning dataset. $\ast$ The evaluation library in MV-GPT \cite{seo2022end} has discrepancy with cococaption \cite{chen2015microsoft} in terms of BLEU4 and METEOR, so the scores are not directly comparable.  ${\ddagger}$ use relationships between sentences to generate paragraphs.}} 
\label{tab:sota}
\end{table*}

\npar{Cross-modal fusion modules.} We conduct ablation studies on cross-modal fusion in Tab.~\ref{tab:crossmodal-fusion}. In terms of the two local fusion modules, merged fusion performs favorably against cross fusion on both datasets. Compared with local fusion modules, global cross fusion performs better on VATEX while it performs worse on YouCook2. 
As shown in Tab.~\ref{tab:language-stats}, the annotated captions are very related to audio events and scenes on VATEX while the captions are largely covered by speech on YouCook2. The results show that global cross fusion helps the flow of holistic information, \eg, acoustic events, while local fusion is more focused on fine-grained information, \eg, single words in speech.
{Moreover, adding global cross fusion greatly improves on both local fusion modules when comparing the results from the first two rows with the last two rows in Tab.~\ref{tab:crossmodal-fusion}.}  

Overall, local-global merged fusion performs best, showing the advantages of using both local and global tokens for cross-modal fusion. 
Since there are only a few global tokens, they do not bring much extra computational cost. Compared with merged fusion, local-global merged fusion increases FLOPs by only 6.96\% (from 316G to 338G), but produces substantially better results.

\npar{Visualization of attention maps.} Fig.~\ref{fig:att_map} shows the maps of the attention from audio modality to the RGB space for global and local fusion modules using Attention Rollout \cite{abnar-zuidema-2020-quantifying}. We visualize the attention maps summed over all the frames in the video clip as in \cite{nagrani2021attention}. 
We note that global fusion focuses on salient regions related to acoustic events (piano), while local fusion is better at leveraging the semantically key words in speech (``\textit{put}'' and ``\textit{spinach}''). 

\begin{figure}
\centering
\vspace{-6mm}
\includegraphics[width=0.85\linewidth]{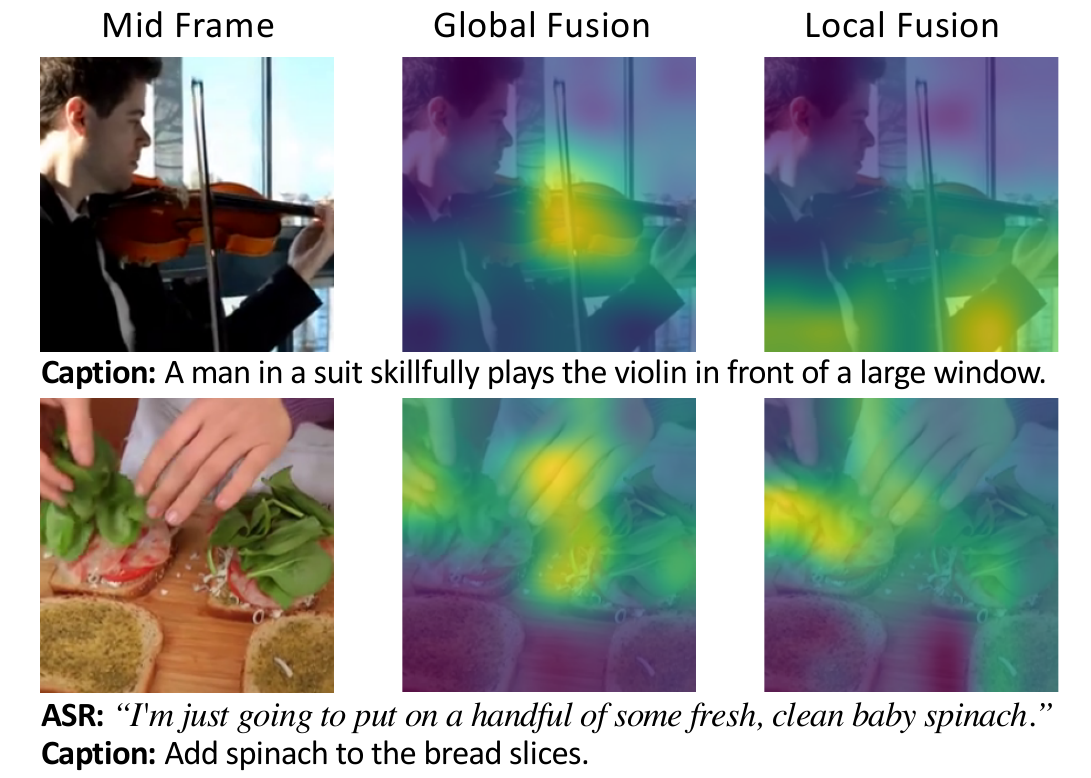}
\vspace{-2mm}
\caption{\small{Attention maps from audio modality to the RGB space for global cross fusion and local fusion modules on VATEX (top) and YouCook2 (bottom).}}
\vspace{-3mm}
\label{fig:att_map}
\end{figure}

\npar{Results with different modalities.} Tab.~\ref{tab:video-audio-text} shows the results when we input different modalities. For audio-only and video-only methods, we zero-mask the other modality at the cross-modal encoder. For V+T setting, we use BERT \cite{kenton2019bert} to replace the audio encoder as in \cite{Luo2020UniVL}. For V+A+T setting, we use global cross fusion between video and audio, and use merged fusion between video and text, to leverage the salient information in audio and text. 
On all four datasets, integrating audio (V+A) leads to a significant improvement compared to the video-only method. 
Besides, comparing V+A with V+T, audio consistently outperforms text. In particular, on VATEX, compared with the video-only method, V+A improves the CIDEr by 6.2\%, while V+T only improves by 1.7\%, showing that audio conveys more information than ASR texts. 
The comparisons between V+A+T and V+T also demonstrate the benefits of incorporating audio for video captioning.
In addition, V+A+T further improves the CIDEr score by 4.6\% on YouCook2 on the basis of the V+A method, probably because the ASR text contains more accurate semantic words than audio. On the other datasets where the speech is not a dominant component in the annotated captions, the performance of V+A+T is very close to that of V+A, showing that audio can cover the major information in speech while bypassing the time-consuming process of ASR \cite{tang2022tvlt}. 
In Fig.~\ref{fig:pipeline-new}, we show two qualitative results on VATEX. In the first video, audio provides more information about laughter and crying, which the ASR text does not contain. In the second video, the V+T method mistakenly recognizes the little girl as a boy and misses the information that a man talks to her, while the V+A method corrects the error and adds the missing description by leveraging the information from audio.

\npar{Comparison with the SOTA methods.} In Tab.~\ref{tab:sota}, we compare the performance of our methods with state-of-the-art methods on four downstream captioning benchmarks. MMCNN \cite{tian2019audio} and MGSA \cite{chen2019motion} are two works that reported audio-visual video captioning results on MSRVTT. Our V+A method clearly outperforms both of them by a large margin. Our proposed V+A framework achieves comparable or even better performance than prior works that rely on information from the text modality, strong pre-trained text encoder/decoder, or additional data for pre-training.
Compared with MV-GPT \cite{seo2022end} that requires ASR transcripts during inference, our audio-visual model generates captions directly from video pixels and audio, bypassing the need of ASR
before generating captions, which is time-consuming \cite{tang2022tvlt}. Contrary to MV-GPT that initializes their model with the weights of BERT \cite{kenton2019bert} and GPT-2 \cite{radford2019language} pre-trained on large-scale document datasets, we train our audio encoder and caption decoder from scratch and do not use any pre-trained language models in our V+A framework. Additionally, the number of parameters of our model (267M) is much smaller than MV-GPT ($>$360M). We prove that directly learning from raw audio and video frames is sufficient for our model to capture key information in the data for generating high quality captions. 
The V+A+T method further improves the performance and achieves the state-of-the-art on YouCook2 and VATEX.
On ActivityNet-Captions, without using any relationships between sentences during training or inference to generate paragraphs, our method can achieve better performance than the methods that exploits relationships across sentences \cite{ging2020coot,pei2019memory}.

\section{Conclusion}
We present an end-to-end pre-training framework for audio-visual video captioning. A novel modality balanced pre-training loss is proposed to balance the learning of different modalities during pre-training, which demonstrates the effectiveness for audio-visual video captioning. We also comprehensively investigate different cross-modal fusion modules for audio-visual fusion and propose a new local-global fusion module. Our model can capture different types of information in human speech and background sounds, achieving comparable or even better results against the models using ASR text or complex language models.

{\small
\bibliographystyle{ieee_fullname}
\bibliography{egbib}
}

\clearpage
\appendix
\renewcommand{\thefigure}{\Alph{figure}}
\setcounter{figure}{0}

In this appendix, we include:
\begin{itemize}
\setlength\itemsep{0em}
    \item Additional implementation details.
    \item Details on audio relevance analysis. 
    \item Additional qualitative results.
    \item Our discussions on potential societal impact.
    \item Our discussions on limitation.
\end{itemize}

\section{Additional Implementation Details}
\label{sec:implementation}
\npar{Paragraph Video Captioning on ActivityNet-Captions}: ActivityNet-Captions contains one reference for each video in the training set and two references for each video in the validation/testing set. Following \cite{zhou2018end}, we use ground-truth segments and sentences for training. For validation/testing, we use the ground-truth segments from the first reference in validation/testing set and evaluate against the two references. 

\npar{Video Encoder}: We equally divide each video clip into $16$ segments and randomly sample $1$ frame from each segment during training. For testing, we uniformly sample 16 frames. Following prior works \cite{lin2022swinbert,liu2022video}, given the raw video frames of the size $T\times H \times W \times 3$, where $T$ is the number of frames, $H \times W$ is the image height and width, and $3$ is the RGB channels, the size of the output features of Video Swin Transformer is $\frac{T}{2}\times \frac{H}{32} \times \frac{W}{32} \times 8C$, where $C$ is the channel dimension. In our experiments, the input size of the video encoder is $16\times 224 \times 224 \times 3$ and the channel dimension is $C=128$, so the output size is $8 \times 7 \times 7 \times 1024$. To be consistent with the token dimension of the other modules, we add a linear layer to map the feature dimension into $768$. Hence, the number of video tokens is $N_v=8\times 7 \times 7 = 392$ and the token dimension is $D=768$.

\npar{Audio Encoder}: Our implementation of the audio encoder is similar to that in \cite{zellers2022merlot}. Each audio is resampled to 22,050Hz and divided into frames of 1536 samples with hop length of $588$. Then we apply 64 mel-scale filters and take logarithm on the amplitude to get the log mel spectrogram. To tackle the variable length of audios, we set a maximum length of frames as 256 ($\sim$6.8 sec.). The audio clips shorter than 256 frames will be zero-padded and longer clips will be down-sampled to be 256 frames. As a result, the dimension of input mel-spectrogram is $256 \times 64$ for each audio. We use a 12-layer Transformer with 12 attention heads on audio spectrogram. We apply the linear 1-dimensional layout on the spectrogram as in \cite{zellers2022merlot} rather than a two-dimensional (image-like) one in \cite{gong2021audio}, as we notice 1-dimensional layout is more suitable for fine-grained audio events like speech. The Transformer produces audio features of dimension $256\times 768$ and then, following \cite{zellers2022merlot},  we apply an average pooling by a factor of four to resize the sequence to a length of $N_a=64$ audio tokens. During training, we use time and frequency masking as in SpecAugment \cite{park2019specaugment} for data augmentation. The time mask parameter is set as $32$ and the frequency mask parameter is set as $128$.

\npar{Cross Encoder}: We use a 3-layer Transformer with 12 attention heads as the cross-modal encoder. The feature embeddings of different modalities will be added with the position embedding and token type embedding to distinguish the position and modality of the tokens.

\begin{figure*}[!t]
    \centering
    \includegraphics[width=0.99\linewidth]{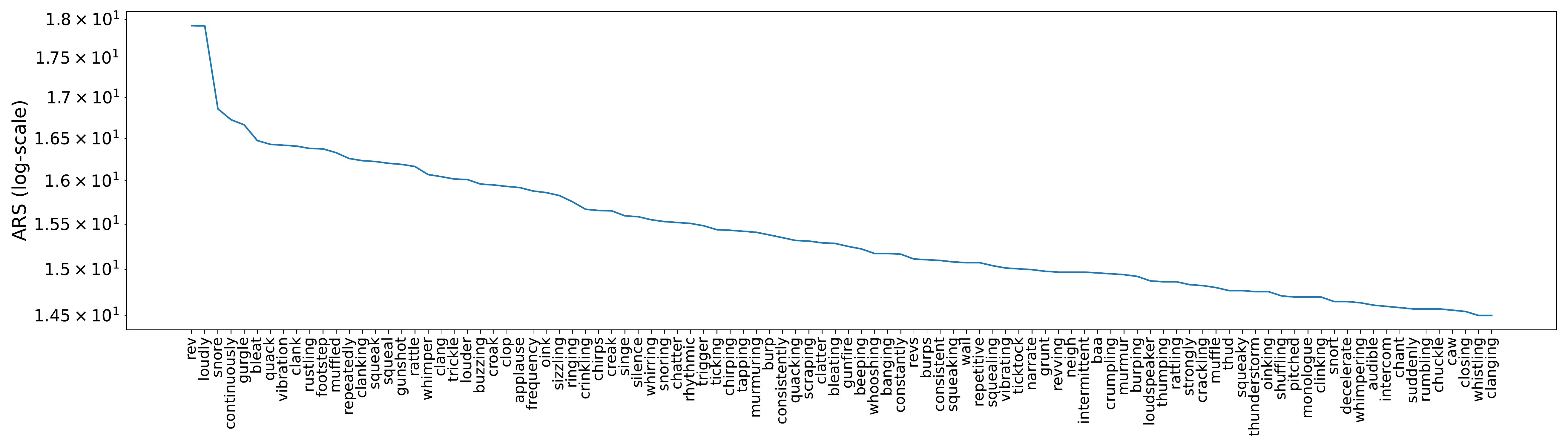}
    \caption {The ARS of top 100 words.}
    \label{fig:ars_top100}
\end{figure*}

\begin{figure*}[tbh]
\centering
  \includegraphics[width=1.0\linewidth]{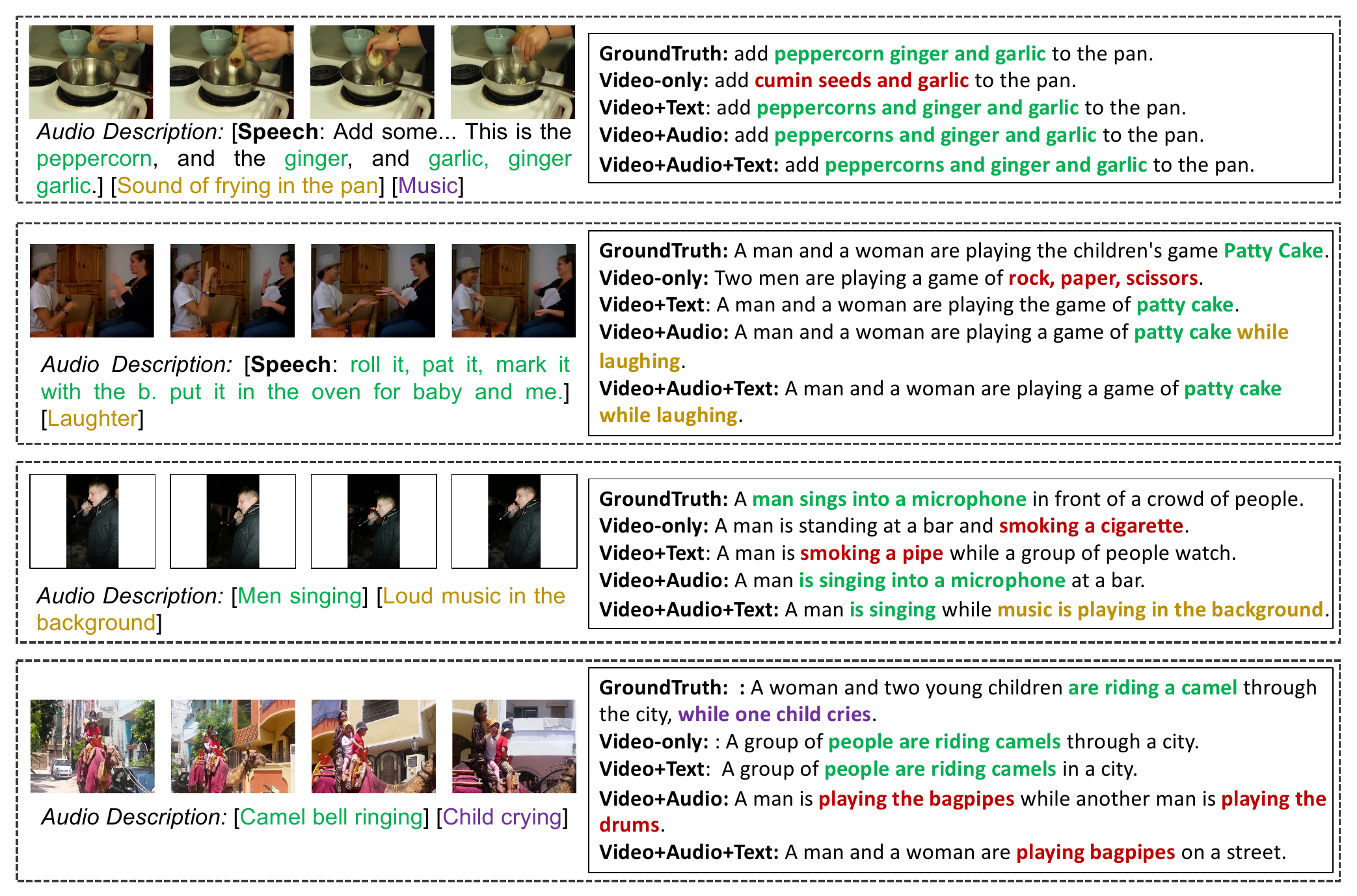}  
  \caption{Some qualitative examples when we input different modalities to generate captions. }
\label{fig:qualitative}
\end{figure*}

\section{Audio Relevance Analysis}
We design two metrics, \ie Speech Coverage Rate (SCR) and Audio Relevance Score (ARS), to measure how relevant the captions are to speech and audio. 

For SCR, we tokenize the annotated captions and speech transcripts, and compute the percentage of tokens in captions that are covered by the associated speech transcripts.

For ARS, we collect the captions from audio captioning and image captioning datasets. We calculate the word frequencies in audio captioning and image captioning datasets, and assign a higher score to the words that occur more frequently in audio captions than image captions. We use the annotated captions from AudioCaps \cite{kim2019audiocaps} and Clotho \cite{drossos2020clotho} as audio captions and the annotated captions from CoCo-Captions \cite{chen2015microsoft} as image captions. We tokenize and lemmatize all captions, and remove all punctuation and stop words. Then we compute the word frequency in audio captions and image captions. Let $f^a(\mathbf{w})$ and $f^i(\mathbf{w})$ denote the frequency of word $\mathbf{w}$ in audio captions and image captions respectively, then ARS is computed as:
\begin{equation}
    \text{ARS}(\mathbf{w}) = \max(\log \frac{f^a(\mathbf{w})}{f^i(\mathbf{w})}, 0).
\end{equation}
Hence, a higher ARS will be assigned to those words whose frequency is much higher in audio captions than that in image captions. On the contrary, if a word is more frequent in image captions, then the log of the division is negative, and the ARS will be set as zero. The ARS of a sentence is the sum of the ARS of all non-stop words in it, and we compute the average ARS of all captions on each dataset.

Figure \ref{fig:ars_top100} shows the ARS of the top 100 words. We can see that most of those words are highly relevant to audio modality, including verbs  that describe a certain type of sound, \eg \textit{``snore", ``gurgle", ``bleat"}, adjectives or adverbs that describe the pattern of sound, \eg \textit{``loudly", ``muffled"}, or nouns that are usually associated with a type of sound, \eg \textit{``vibration", ``gunshot"}, \etc.

\section{Additional Qualitative Results}
\label{sec:exp}

\spar{Qualitative results.} 
We show some qualitative examples in Fig.~\ref{fig:qualitative} by comparing the ground truth, and predictions of our model with video-only, video+text, video+audio, and video+audio+text inputs. In the first example, when only visual modality is available, the model misclassifies the peppercorns as cumin seeds as they are hardly visible in the video. By adding audio and/or ASR text as input, the model correctly generates the caption because the ingredients are clearly introduced in the speech, which shows that we can infer the ingredient from raw audio without relying on an off-the-shelf ASR system. In the second example, the caption is not explicitly described in the audio, but the model is able to correct its prediction from ``rock, paper, scissors" to ``patty cake" by adding audio or text modality as it can infer the type of game from the speech. In particular, when we add audio as the input, the model not only predicts the correct type of game, but also provides more information about the audio, \ie recognizing that the people are laughing. The third example shows that audio is also helpful when there is no speech in the video. Using video-only or video+text as the inputs, the model generates incorrect captions, \eg ``smoking a cigarette" or ``smoking a pipe" as the microphone is not quite visible. However, if we add audio as the input, the model will correctly detect that the man is singing and music is playing in the background. The last example is a failure case, where the audio-visual model mistakenly recognizes the sounds of camel bell ringing and child crying as the sounds of the bagpipes and the drums. It indicates that the model gives too much attention to the audio modality and ignores the visual appearance of people riding a camel. 

\begin{figure}
\centering
\vspace{-6mm}
\includegraphics[width=0.85\linewidth]{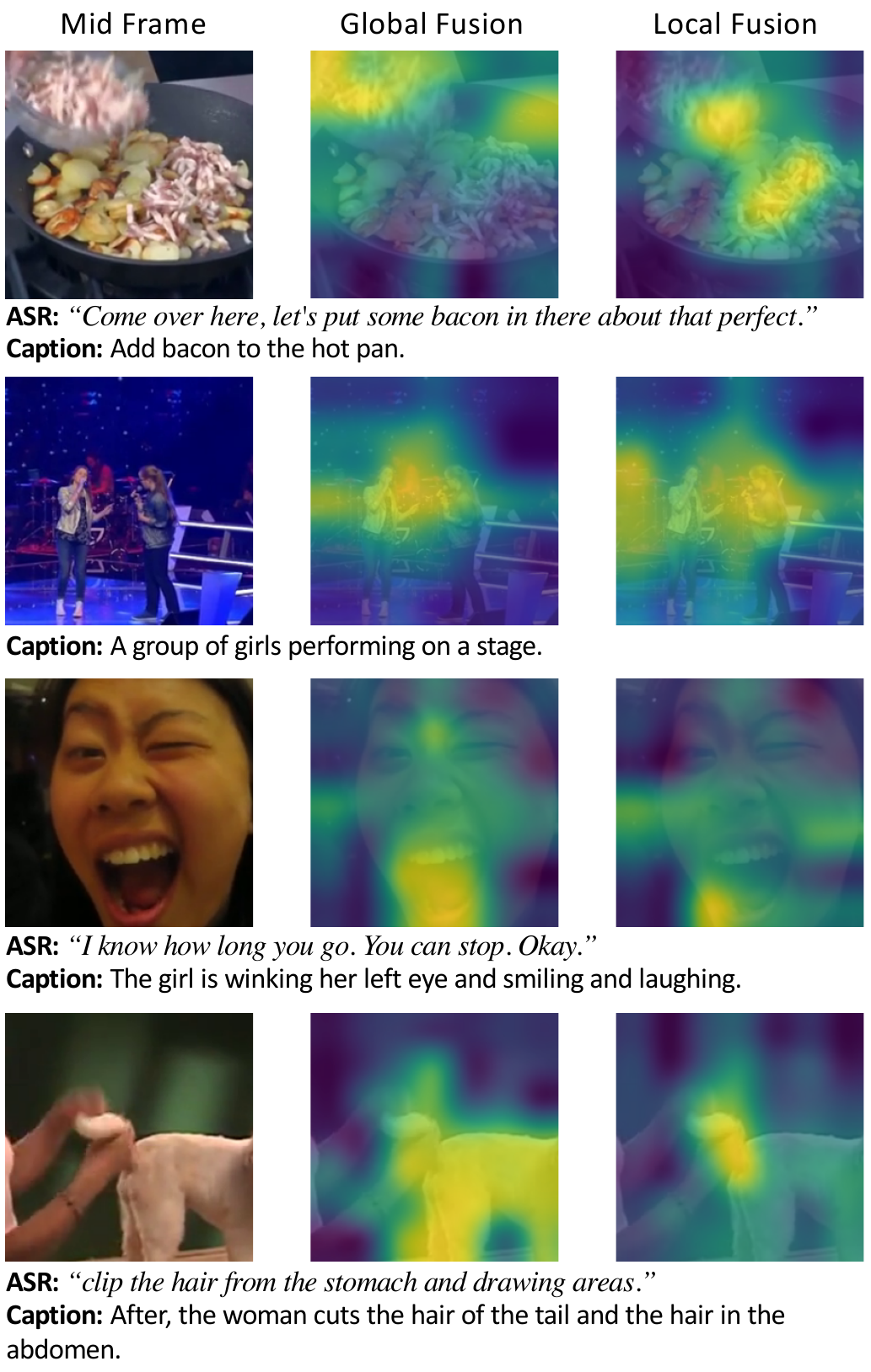}
\vspace{-2mm}
\caption{\small{Additional visualization on the attention maps from audio modality to the RGB space for global cross fusion and local fusion modules on YouCook2, MSRVTT, VATEX, and ActivityNet, respectively.}}
\vspace{-3mm}
\label{fig:att_map}
\end{figure}

\spar{Attention maps.} We show additional visualization on the attention maps on the four downstream datasets respectively. Similar to what we observed in the main paper, in the first and the fourth example, where the items in the videos are mentioned in the speech, \ie, ``\textit{bacon}" and ``\textit{hair}", the local attention will focus on the regions of those items. In the second example, where there is sound of singing in the video, both local and global modules will give attention to the performers on the stage, and we notice that the attention of the global module will be more concentrated. In the third example, where there is sound of speaking and laughter, the global module focuses more on the mouth region, where the sound comes from.

\section{Discussions on Societal Impact}
\label{sec:impact}
Video captioning makes videos more accessible to all users, including users with accessibility issues, \eg lowvision and blind users. Furthermore, audio-visual video captioning also benefits users with hearing disability by including text descriptions of the audio modality. However, our framework is a data-driven system, so the quality of generated captions may be biased to the distribution of training data. As our pre-training data are obtained from online YouTube videos, our system may produce harmful captions that contains toxic contents or social biases present in the training data. To avoid undesirable effects, careful examination is required before adopting the outputs of our framework for real-world applications. Another ethical concern on using YouTube videos is how to protect user privacy. In our experiments, we download the videos currently available on YouTube. Therefore, if a user deletes a video from their YouTube channel, we will not be able to use the videos.

\section{Discussions on Limitations}
\label{sec:limit}

The failure case in Fig.~\ref{fig:qualitative} shows that our approach is not always successful in the fusion of visual and audio modalities. A potential direction is to design a more intelligent cross-modal fusion module to dynamically update the weights of different modalities for caption generation.
Besides, the dominance of the modality is dataset-dependent, and we pre-trained our audio-visual video captioning model on the HowTo100M dataset, where audio dominates due to the use of ASR transcripts as text supervision. We did not verify the effectiveness of our proposed modality balancing pre-training strategy on vision-dominant datasets due to the lack of such large-scale pre-training datasets with video-text pairs. We believe that our proposed approach has the potential to be advantageous in those settings, and it could be explored in future research.

\end{document}